\documentclass{article}

\usepackage[preprint]{neurips_2026}

\usepackage[utf8]{inputenc}
\usepackage[T1]{fontenc}
\usepackage{algorithm}
\usepackage{algpseudocode}
\usepackage{hyperref}
\usepackage{url}
\usepackage{booktabs}
\usepackage{amsmath}
\usepackage{amsfonts}
\usepackage{graphicx}
\usepackage{float}
\usepackage{xcolor}

\title{Weight Pair Encoding: Inducing a Smaller Grammar in Neural Network Weights}

\newcommand{\authorbox}[1]{\begin{minipage}[t]{0.46\linewidth}\centering\mdseries #1\end{minipage}}
\author{%
  \authorbox{\textbf{Irene Tallini}\thanks{Equal contribution.}\\
  Area Science Park\\
  Trieste, Italy\\
  \texttt{irene.tallini@areasciencepark.it}}
  \And
  \authorbox{\textbf{Daniele Solombrino}\footnotemark[1]\\
  Sapienza University of Rome\\
  Rome, Italy\\
  \texttt{solombrino@di.uniroma1.it}}
  \And
  \authorbox{\textbf{Alberto Cazzaniga}\thanks{Equal senior contribution.}\\
  Universit\'e C\^{o}te d'Azur, Inria, LJAD\\
  Maasai Project Team, Nice, France\\
  Area Science Park, Trieste, Italy}
  \And
  \authorbox{\textbf{Emanuele Rodol\`a}\footnotemark[2]\\
  Sapienza University of Rome and Paradigma\\
  Rome, Italy}
}

\begin{document}

\maketitle

\begin{abstract}
We show that neural network weights can be explicilty fintuned to admit a smaller grammar. Weight Pair Encoding (WeightPE) does so by placing a lossy Re-Pair compressor
inside a straight-through estimator. The int8 weights of the network are flattened into one string, and near-matching Re-Pair patterns are made exactly equal within a global $L_2$ budget. The network computes with the rewritten weights and trains through them with a straight-through estimator.
Unlike a flat codebook of fixed-size entries, a grammar offers variable-length patterns and reuses them hierarchically inside larger ones.
On the MLP weights of ViT-B/16 and ViT-L/16 finetuned on CIFAR-10,
WeightPE produces a Re-Pair grammar 0.43× and 0.38× the size of the one produced by an equivalent int8 QAT run, at a cost of 1.9 and 1.1 accuracy points. The trend extends to different grammar compressors (LZ78, SEQUITUR), over which the networks has not be finetuned against. To our knowledge, this is the first time grammar size has been used as an explicit training objective for network weights.
\end{abstract}

\section{Introduction and Related Works}
Identifying and imposing structure in the weights of a network has repeatedly proved valuable: weight repetition \citep{nowlan1992soft,ullrich2017soft}, low-rank structure \citep{aghajanyan2021intrinsic,hu2022lora}, permutation symmetry \citep{entezari2022role,ainsworth2023gitrebasin}.

Grammar structure is another rich candidate: it compresses a string \textit{exposing its repetitions hierarchically}, it supports random access in space proportional to the compressed size \citep{bille2015random} and direct computation over the compressed representation \citep{ferragina2022matvec}. The size of the smallest grammar generating a string is an expressive yet computable approximation of Kolmogorov complexity: finding it is NP-hard, but it admits heuristics with worst-case guarantees \citep{charikar2005}.

In principle, any scheme that reduces the symbol alphabet shrinks the grammar: quantization, by leaving fewer distinct symbols, leaves fewer distinct substrings, and the grammar shrinks as a side effect. Existing compact representations of weights work this way and are flat: codebooks and clustering \citep{han2016deepcompression,cho2022dkm}, or size placed directly in the objective through entropy penalties, learned bit-widths, and reusable motifs \citep{oktay2020scalable,csefalvay2023selfcompressing,bakhtiarifard2026momos}. The atom is always one weight or a fixed-size block, so variable-length substrings that recur, and recur inside one another, cannot be named. To our knowledge, grammar size has never been an explicit training objective for network weights.

We therefore introduce \textbf{WeightPE}, which \textbf{pushes the weights at training time to admit a small grammar}. WeightPE flattens the int8-quantized weights into a single string and compresses it with a lossy variant of Re-Pair: recurring patterns are first rewritten onto shared cluster leaders under a global L2 distortion budget and standard Re-Pair merges then build the grammar, similarly to a lossy tokenization of the flattened, concatenated weight matrix. The gradient is computed through STE. On ViT-B/16 and ViT-L/16 finetuned on CIFAR-10, the MLP weights \textbf{WeightPE produces a Re-Pair grammar $0.43\times$ and $0.38\times$ the size of the one an equivalent int8 QAT run, at a cost of $1.9$ and $1.1$ accuracy points.}

\begin{figure}[!t]
  \centering
  \includegraphics[
    width=0.75\linewidth,
    trim=0 0 95mm 0,
    clip
  ]{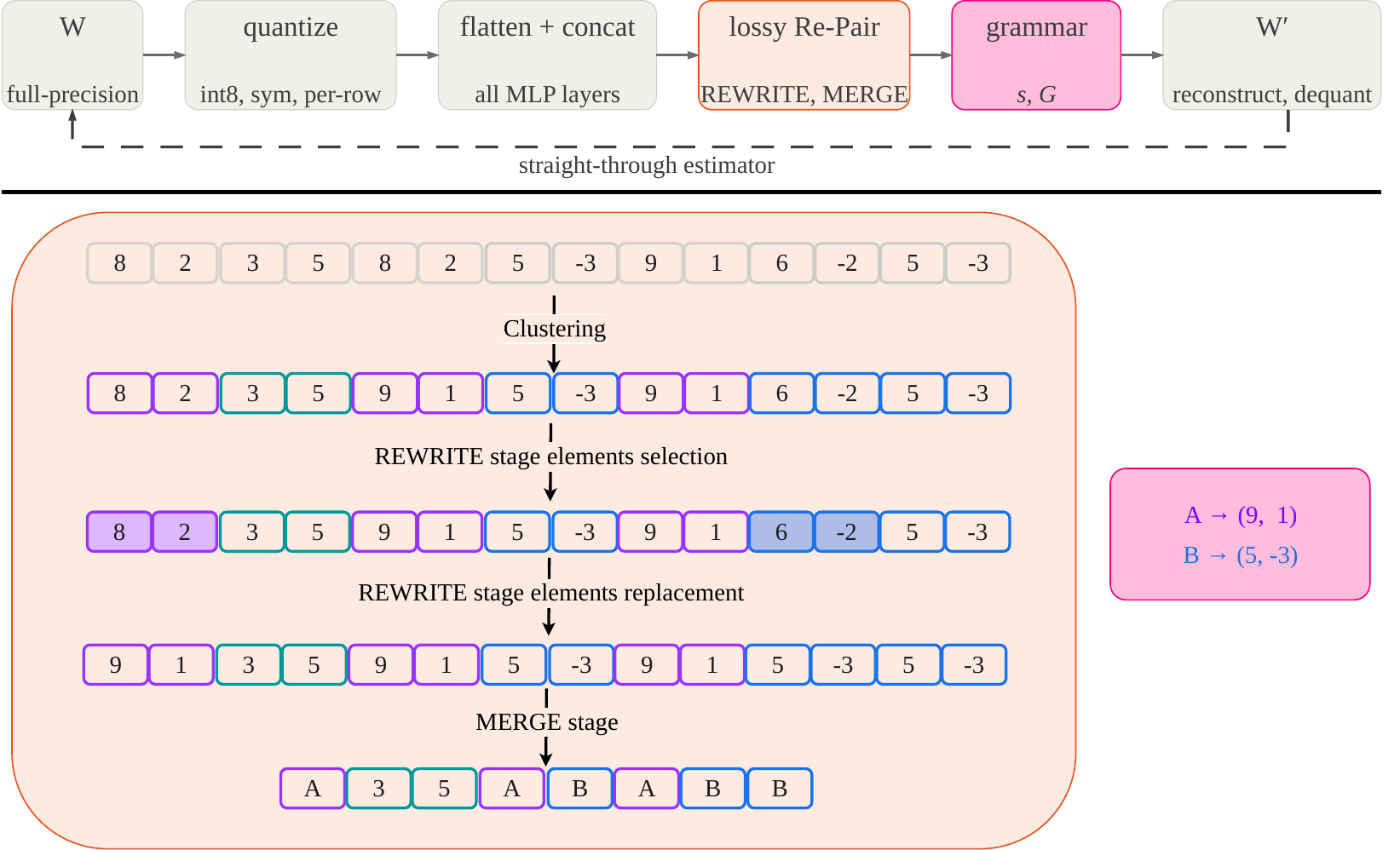}
  \caption{\textbf{Top:} Finetuning pipeline of WeightPE.
  \textbf{Bottom left:} High-level steps of one lossy Re-Pair iteration.
  \textbf{Bottom right:} Cluster leaders emerging in one step of WeightPE.}
  \label{fig:method}
\end{figure}

\section{Background}
\label{sec:background}

\paragraph{Re-Pair}
Re-Pair \citep{larsson2000repair}, the compression algorithm byte-pair encoding
\citep{gage1994bpe, sennrich2016bpe} is built upon, approximates the smallest grammar by
repeatedly replacing the most frequent adjacent pair of symbols with a new
non-terminal, until no pair is present twice. Its rules are all couples, so a Re-Pair grammar with rule set $G$ over a residual sequence $s$ has size
$|s| + 2|G|$ symbols: the measure of \citet{charikar2005}, which we report
throughout.

\paragraph{Straight-through estimation}
Quantization-aware training \citep{jacob2018quantization} computes the forward
pass on quantized weights, avoiding the non-differentiability issues by treating the displacement the map introduces as a constant rather than as a function of the weights \citep{bengio2013ste}. For a map $P$ applied to a weight matrix $W$,
let $c = P(W) - W$, held fixed with respect to $W$; the network then computes
with
\begin{equation}\label{eq:ste}
  W' \;=\; W + c \;=\; P(W),
\end{equation}
while the gradient reaches $W$ unchanged, as if $P$ were the identity.

\section{Method}
\label{sec:method}

Nothing in \eqref{eq:ste} requires $P$ to be a quantizer: it can be any operator which does not irreparaiably degrade the loss.
We build a $P$ that also makes the grammar of the weights small. It quantizes, serializes the whole network into one string, and compresses that string with a lossy Re-Pair that may merge patterns close in $L_2$ distance (Figure~\ref{fig:method})\footnote{We note the problem can be formulated as a rate--distortion problem.}.

\paragraph{Serialization}
Each weight matrix is quantized to int8 row-wise \citep{krishnamoorthi2018whitepaper}, flattened row-major, and concatenated into a single string.
One grammar therefore covers the whole network, and a pattern found in one block is reusable in another.
We also impose that no rule can cross a row.

\paragraph{Lossy Re-Pair}
Re-Pair shares a rule between two substrings only when they are exactly equal, whereas a quantized weight matrix is full of patterns that are \textit{almost} equal. Our operator therefore
alternates two rounds. \textsc{REWRITE}, the only lossy step, buckets
adjacent-pair occurrences by the length of the code string they expand to;
within a bucket the $T$ most frequent distinct pairs become cluster leaders,
and each occurrence is rewritten onto its nearest leader, cheapest first according to \(L_2\) distance over a
non-overlapping set, until the distortion budget is spent. \textsc{Merge} is a
lossless Re-Pair round on the rewritten sequence, so the near-duplicates that
\textsc{REWRITE} has just made exactly equal collapse into one shared rule. Unlike
standard Re-Pair it merges every repeated pair rather than only the most
frequent, for computational efficiency. Once the spent distortion reaches the budget,
\textsc{REWRITE} switches off and the remaining rounds are plain Re-Pair.
Please refer to Appendix~\ref{app:operator} for further details about the algorithm.

\paragraph{Training}
The perturbed string is expanded back to weights $W'$ and trained through \eqref{eq:ste}.
For computational efficiency, the perturbation operator is applied every ten optimization steps.

\begin{figure}[!t]
  \centering

  \includegraphics[width=\linewidth]{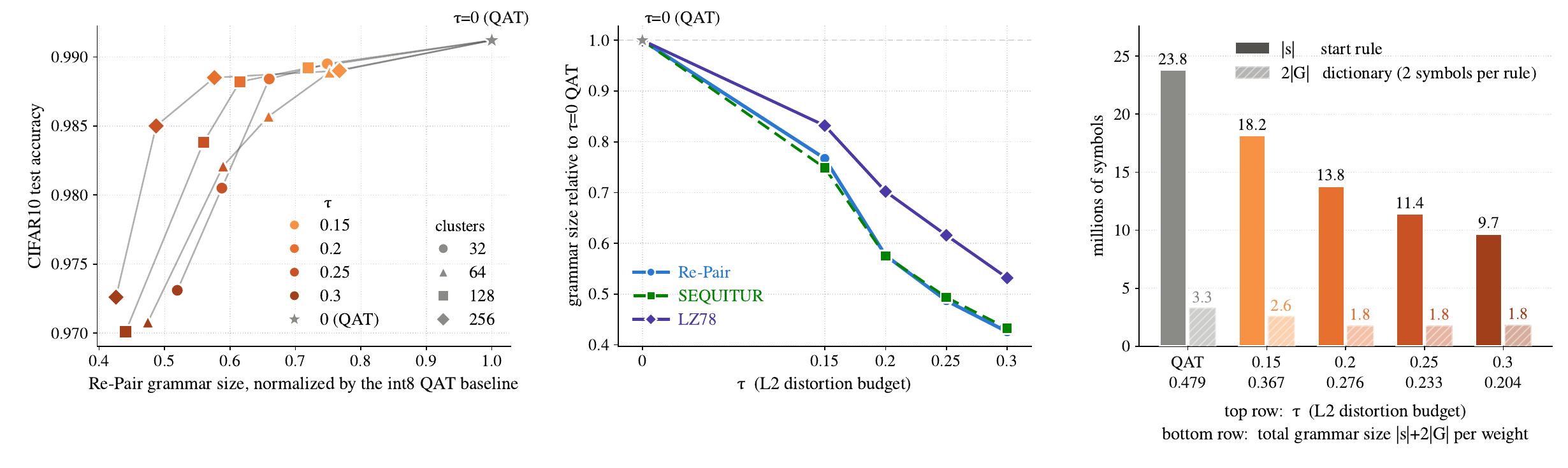}
  \vspace{-0.5em}

  \includegraphics[width=\linewidth]{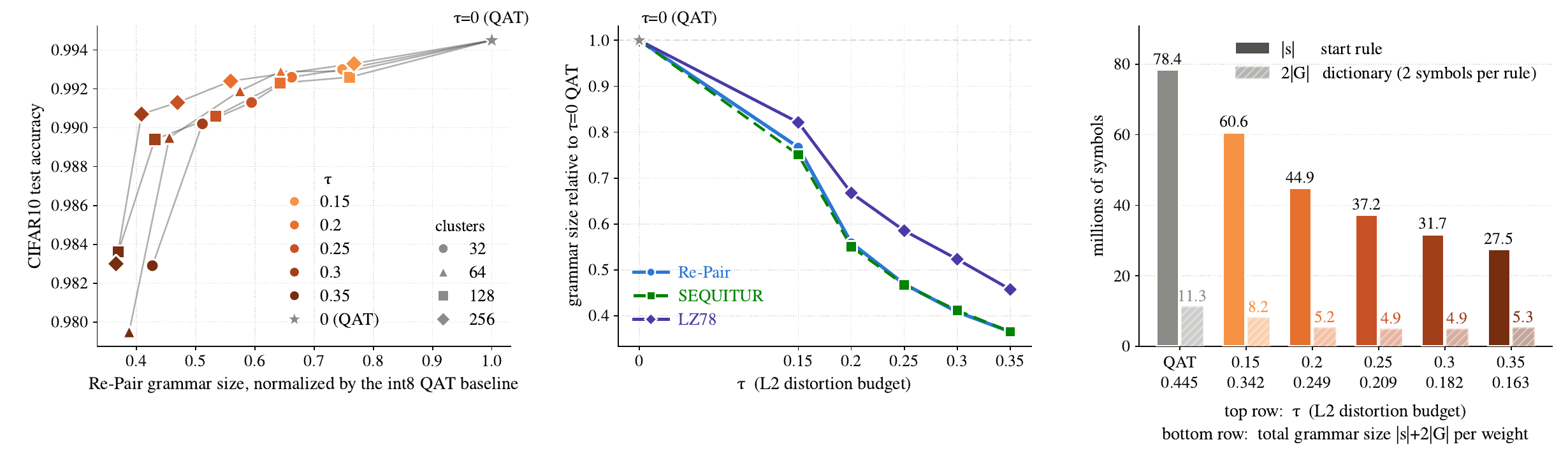}

  \caption{WeightPE test accuracy and grammar size for ViT-B/16 (top) and ViT-L/16 (bottom) on CIFAR-10.
  \textbf{Left:} the effect of grammar size, normalized by the baseline grammar size, on test accuracy, for various numbers of cluster leaders \(T\) and budget errors \(\tau\).
  \textbf{Center:} the effect of \(\tau\) on grammar size under different compressors.
  \textbf{Right:} the effect of \(\tau\) on absolute grammar size.
  }
  \label{fig:tradeoff}
\end{figure}

\section{Experiments}
\label{sec:experiments}

We finetune ViT-B/16 and ViT-L/16 on CIFAR-10 (Appendix~\ref{app:setup}).
The operator applies to any weight matrix, but we restrict the string to the two MLP projections of every block, which carry about two thirds of the parameters.
The attention block passes through a softmax over its own activations, so a perturbation of its weights propagates less predictably.
The attention projections are quantized to int8 through the same estimator but never perturbed, while patch embedding, normalization, and classification layers are left in full precision~\citep{or2025torchao}. We sweep the distortion budget $\tau$ and the number of cluster leaders $T$.

\paragraph{Metrics}
We use top-1 accuracy on the CIFAR-10 test set as downstream task evaluation metric.
The smallest grammar of the deployed weights cannot be computed: each compressor only gives an upper bound on it.
Thus we report three grammar compressors: Re-Pair, which the operator is built against; SEQUITUR \citep{nevillmanning1997sequitur} and LZ78 \citep{ziv1978lz78}, which it is not.
All grammar sizes are normalized by the corresponding grammar size of the $\tau{=}0$ int8 QAT baseline, trained under identical settings.
Both quantities consider only the layers used in the string (MLP weights), as reported in~\ref{sec:experiments}.

\section{Results}
\label{sec:results}

On ViT-B/16 the Re-Pair grammar of the deployed weights reaches $0.43\times$ the size of the QAT grammar for $1.9$ accuracy points; on ViT-L/16, $0.37\times$ for
$1.1$.

\paragraph{Accuracy vs. grammar size tradeoff}
The left column of Figure~\ref{fig:tradeoff} plots accuracy against grammar size, showing one curve for each cluster count \(T\), at various $\tau$ values.
On ViT-B/16 the grammar size falls down to $0.49\times$ when compared to the baseline int8 QAT run, losing just $0.7$ points of downstream test accuracy.
Further grammar size reductions down to $0.43\times$ reach $97.3\%$ test accuracy.
The number of leaders matters only where the budget is large enough not to be spent all on the short sequences.
At $\tau{=}0.15$ the four curves sit on top of one another, while at $\tau{=}0.30$ they separate cleanly.
The right column gives an indication about absolute sizes involved in the ratios and shows that much of the compression comes from a length reduction of the top-level string.
Wall execution time, reported in Figure~\ref{fig:timing-matrix}, grows with both WeightPE hyperparameters.

\paragraph{Generalization across compressors}
The central column shows the grammar size obtained when using three compressors on the same weights.
Each grammar size is normalized by the grammar size of the baseline int8 QAT run.
SEQUITUR matches Re-Pair rates almost perfectly, while LZ78 moves the least.
Notably, both of these compressors have never been used in the WeightPE perturbation applied during finetuning.

\paragraph{Scaling model size}
ViT-L/16 exhibits a more favorable accuracy--grammar-size trade-off than ViT-B/16.
At a comparable grammar size of approximately \(0.38\times\) the int8 QAT baseline, ViT-L/16 incurs only a \(0.3\)-percentage-point drop in test accuracy, compared with a \(1.9\)-point drop for ViT-B/16.
Moreover, ViT-L/16 remains robust as the distortion budget is increased further: at the largest tested \(\tau\), the accuracy degradation is only \(1.1\) percentage points, compared with approximately \(1.9\) points for ViT-B/16 at its strongest compression setting.
These results suggest that increasing model scale substantially improves robustness to grammar compression, enabling more aggressive compression at a given downstream accuracy loss.

\section{Conclusion}
\label{sec:conclusion}

We introduced Weight Pair Encoding, which makes the grammar size of a network's weights an explicit
training objective by placing a lossy Re-Pair perturbation, bounded by a global $L_2$ budget,
inside a straight-through estimator. On two ViT backbones finetuned on CIFAR-10 the deployed
weights admit a grammar roughly $2.5\times$ smaller than the one a parallel int8 QAT at a small cost in accuracy, and the reduction is visible to compressors the objective
was not built from.

\paragraph{Limitations}
Behavior for more model families, datasets and quantization levels still has to be tested. We swept only row-major flattening, although the serialization axis is a free choice of the method and a different one exposes a different set of patterns. Finally, the operator has not yet been optimized for computatoonal efficiency: its cost is negligible for some parameters and around 5x in others.

\paragraph{Future work}
The perturbation is built on Re-Pair, but the same idea can be in principle applied to any compressor with an explicit notion of a repeated pattern. It would be interesting to optimize for good permutation first, like \citet{ainsworth2023gitrebasin}, and then apply WeightPE. Since robustness to quantization has been shown to be a transferable direction in weight space \citep{solombrino2026zeroshot}, it is natural to ask whether the small-grammar structure induced by WeightPE is one too, so that a model could be patched toward a smaller grammar without finetuning. It is also open whether a shared rule corresponds to an interpretable computation, whether a grammar over weights helps the models that consume weights as data, and how the scheme compares to standard compressors once size is counted in bits rather than in symbols.

\section*{Acknowledgments}
The authors acknowledge the Area Science Park supercomputing platform ORFEO made available for
conducting the research reported in this paper, and the technical support of the Laboratory of
Data Engineering staff.
IT and AC were supported by the projects ``Supporto alla diagnosi di malattie rare tramite
l'intelligenza artificiale'', CUP: F53C22001770002, and ``Valutazione automatica delle immagini
diagnostiche tramite l'intelligenza artificiale'', CUP: F53C22001780002. AC was supported by the
European Union -- NextGenerationEU within the project PNRR ``PRP@CERIC'' IR0000028 -- Mission 4
Component 2 Investment 3.1 Action 3.1.1. This work has been carried out while DS was enrolled in the Italian National Doctorate on Artificial Intelligence run by Sapienza University of Rome.

\vfill
\bibliographystyle{plainnat}
\bibliography{bib}

\appendix

\newpage

\section{The operator in detail}
\label{app:operator}

\paragraph{All merges together to decrease runtime.}
Standard Re-Pair performs one merge (the single globally most frequent pair) per step, which
is unfeasible at the scales of our experiments. Our operator instead collapses a whole \emph{independent set} of non-overlapping
occurrences per round, all in GPU, so the number of
sequential rounds is $\mathcal{O}(\log N)$ rather than $\mathcal{O}(N)$ and the weight string never
leaves the GPU. The grammar this produces is therefore not the one sequential greedy Re-Pair
would produce: the sequential implementation is only used for evaluation.
Algorithms~\ref{alg:perturb}--\ref{alg:merge} give the whole procedure. $x$ is the sentence of int8 codes, $s$ the current symbol sequence, $G$ the rule set, $\mathit{pos}_j$ the offset at which symbol $s_j$ expands in $x$, and $r$ the running exact reconstruction of $x$ under the current grammar.

\begin{algorithm}[t]
\caption{The perturbation operator: REWRITE, then merge, once per round.}
\label{alg:perturb}
\begin{algorithmic}[1]
\Function{Perturb}{$x$, $\tau$, $n_{\max}$}
  \State $s \gets x$; \enspace $G \gets \emptyset$; \enspace $\mathit{pos}_i \gets i$;
         \enspace $r \gets x$; \enspace $D \gets 0$; \enspace $\mathit{stall} \gets 0$
  \Loop
    \State $b \gets |s| + 2|G|$
    \If{$b \le n_{\max}$ \textbf{ or } $|s| < 2$} \textbf{break}
      \Comment{(i) the grammar reached its target size} \EndIf
    \State $n_{\mathrm{REWRITE}} \gets 0$; \enspace $\delta \gets 0$
    \If{$D < \tau^2$} \Comment{budget exhausted $\Rightarrow$ plain Re-Pair from here on}
      \State $(s, \mathit{pos}, r, n_{\mathrm{REWRITE}}, \delta) \gets
             \Call{REWRITERound}{s, \mathit{pos}, r, x, \tau^2 - D}$; \enspace $D \gets D + \delta$
    \EndIf
    \State $(s, \mathit{pos}, G, n_{\mathrm{new}}) \gets \Call{MergeRound}{s, \mathit{pos}, G}$
    \If{$n_{\mathrm{REWRITE}} = 0$ \textbf{ and } $n_{\mathrm{new}} = 0$} \textbf{break}
      \Comment{(ii) nothing rewritten and nothing merged} \EndIf
    \If{$|s| + 2|G| \ge b$}
      \State $\mathit{stall} \gets \mathit{stall} + 1$;
             \textbf{ if } $\mathit{stall} = 2$ \textbf{ then break}
             \Comment{(iii) stalled, after one grace round}
    \Else \State $\mathit{stall} \gets 0$
    \EndIf
  \EndLoop
  \State \Return $s$, $G$, $D$
\EndFunction
\end{algorithmic}
\end{algorithm}

\begin{algorithm}[t]
\caption{The lossy step. $B$ is the distortion still affordable, $\rho = \tau/\sqrt{N}$ the leader
de-duplication radius per element, $T$ the number of clusters.}
\label{alg:REWRITE}
\begin{algorithmic}[1]
\Function{REWRITERound}{$s$, $\mathit{pos}$, $r$, $x$, $B$}
  \State $r_0 \gets r$ \Comment{round-start snapshot: REWRITEs do not see each other's REWRITEs}
  \State $\mathrm{cost}_j \gets \infty$ for every occurrence $j$, i.e.\ every pair $(s_j, s_{j+1})$
         lying inside one row
  \ForAll{length buckets $L \ge \ell_{\min}$ holding $k \ge c_{\min}$ occurrences}
    \State $V \gets \Call{Leaders}{\text{distinct pairs of the bucket}, T, \rho, r_0}$
    \ForAll{occurrences $j$ of the bucket}
      \State $\chi \gets x[\mathit{pos}_j : \mathit{pos}_j + L]$
             \Comment{the immutable original codes of this span}
      \State $e \gets \lVert r_0[\mathit{pos}_j : \mathit{pos}_j + L] - \chi \rVert^2$
             \Comment{what the span already costs}
      \State $t^{\ast} \gets \arg\min_t \lVert V_t - \chi \rVert^2$,
             over the leaders whose pair differs from $j$'s
      \State $\mathrm{cost}_j \gets \lVert V_{t^{\ast}} - \chi \rVert^2 - e$; \enspace
             $\mathrm{leader}_j \gets t^{\ast}$
    \EndFor
  \EndFor
  \State $S \gets \Call{SelectIndependent}{-\mathrm{cost}, \{\, j : \mathrm{cost}_j < \infty \,\}}$
  \State sort $S$ by increasing cost, keep the longest prefix whose cumulative cost is $\le B$
         \Comment{the budget, enforced a priori}
  \ForAll{$j \in S$}
    \State overwrite $s_j, s_{j+1}$ with the two symbols of $\mathrm{leader}_j$
    \State $\mathit{pos}_{j+1} \gets \mathit{pos}_j + |\text{its left symbol}|$
           \Comment{the leader's split need not match $j$'s own}
    \State $r[\mathit{pos}_j : \mathit{pos}_j + L] \gets$ the leader's values, read from $r_0$
  \EndFor
  \State \Return $s$, $\mathit{pos}$, $r$, $|S|$, $\sum_{j \in S} \mathrm{cost}_j$
\EndFunction
\Statex
\Function{Leaders}{$P$, $T$, $\rho$, $r_0$} \Comment{$P$: the bucket's distinct pairs, with counts}
  \State $A \gets \langle\,\rangle$
  \ForAll{$p \in P$ in decreasing frequency, at most $8T$ candidates}
    \If{$A = \langle\,\rangle$ \textbf{ or }
        $\min_{a \in A} \lVert \mathrm{val}(p) - \mathrm{val}(a) \rVert^2 > \rho^2 L$}
      \State append $p$ to $A$; \textbf{ if } $|A| = T$ \textbf{ then break}
    \EndIf
  \EndFor
  \State \Return the expansions of $A$, read from $r_0$
\EndFunction
\end{algorithmic}
\end{algorithm}

\begin{algorithm}[t]
\caption{The lossless step, and the selection rule both steps share.}
\label{alg:merge}
\begin{algorithmic}[1]
\Function{MergeRound}{$s$, $\mathit{pos}$, $G$}
  \State $C \gets \{\, j : (s_j, s_{j+1}) \text{ lies inside one row and occurs at least twice} \,\}$
  \State $S \gets \Call{SelectIndependent}{\mathrm{freq} + \mathrm{jitter}, C}$
         \Comment{the jitter only makes the order strict}
  \State add one fresh rule $A \rightarrow (s_j, s_{j+1})$ to $G$ per distinct pair occurring in $S$
  \State replace every $j \in S$ by its rule symbol and delete the consumed right symbol
         ($\mathit{pos}_j$ is inherited)
  \State \Return $s$, $\mathit{pos}$, $G$, the number of new rules
\EndFunction
\Statex
\Function{SelectIndependent}{$\pi$, $C$}
  \State \Return $\{\, j \in C : \pi_j > \pi_{j-1} \text{ and } \pi_j > \pi_{j+1} \,\}$
         \Comment{at the ends, the missing neighbour counts as beaten}
\EndFunction
\end{algorithmic}
\end{algorithm}

\paragraph{A worked example}
The following is one round of \textsc{Perturb} on a twelve-code sentence, run through the
implementation. The budget is $\tau^{2} = 5$ and the bucket gets $T = 2$ leaders. Every symbol is
still a single code, so occurrence $j$ covers positions $j$ and $j+1$, the reconstruction equals
the original, and each cost is simply the squared distance from the chosen leader.

{\small
\begin{verbatim}
position      0   1   2   3   4   5   6   7   8   9  10  11
x           = 5   3   5   5   3   5   8   2   5   3   4   6

occurrence j    0     1     2     3     4     5     6     7     8     9    10
pair          (5,3) (3,5) (5,5) (5,3) (3,5) (5,8) (8,2) (2,5) (5,3) (3,4) (4,6)
cost            8     8     4     8     8    13    10     1     8     1     2

leaders (the two most frequent pairs of the bucket):  (5,3) x3,  (3,5) x2
independent set (strict local minima of the cost):    j2, j7, j9
sorted by cost: 1, 1, 4     cumulative: 1, 2, 6     B = 5  =>  commit j7 and j9

after REWRITE  = 5   3   5   5   3   5   8  [3   5] [3   5]  6        D = 2
\end{verbatim}}

\noindent
$j_2$ does not fit in the budget and stays a candidate for the next round. The two committed REWRITEs
cost one each: $(2,5) \rightarrow (3,5)$ and $(3,4) \rightarrow (3,5)$. Their point is what happens
next: the pair $(3,5)$ now occurs four times instead of two, at $j_1$, $j_4$, $j_7$ and $j_9$, which
are pairwise non-adjacent, so the \textsc{MergeRound} that follows creates a single rule covering
all four. The round returns $|s| = 8$ with one rule, a grammar of $10$ symbols against the $12$ it
started from.

\section{Experimental setup}
\label{app:setup}

\paragraph{Models and data}
We finetune \texttt{timm} \citep{wightman2019timm} ViT-B/16 and ViT-L/1 checkpoints \citep{dosovitskiy2021vit} pretrained on ImageNet-21k with the AugReg recipe \citep{steiner2022augreg}
on CIFAR-10 \citep{krizhevsky2009cifar}, with a freshly initialized $10$-way head; input transforms
are taken from each checkpoint's own pretraining configuration, and the validation split is drawn
with a dedicated seed independent of the run seed.

Following \cite{ilharco2022editing}, we finetune for six epochs, with an effective batch of $128$ reached by gradient accumulation, AdamW at learning rate $10^{-5}$ with weight decay $0.1$, a cosine schedule with $500$ warm-up
steps, gradient clipping at $1.0$, and no label smoothing.
Every run in each sweep share these hyperparameters, so runs differ only in $\tau$ and in the cluster count $T$.

\paragraph{WeightPE}
We Apply WeightPE to the two MLP projections of every transformer block, corresponding to almost two thirds of the total number of trainable parameters.
The query, key, and value projections are quantized, but never perturbed.
Patch embedding, normalization, and classification layers are kept in full precision, following \citep{or2025torchao}
We use row-wise, symmetric, int8 weights-only quantization.
Accordingly, we adopt row-major serialization and we confine rules to a row, while maintaining a single grammar shared across all perturbed MLP weights in the network.
We set the minimum expansion length for \textsc{Rewrite} to $\ell_{\min}=2$ and require at least $c_{\min}=2$ occurrences in a length bucket before processing it.
The perturbation is run with target grammar size $n_{\max}=64$.
For leader construction, we use a de-duplication radius of $\rho=\tau/\sqrt{N}$, where $N$ is the number of quantized codes in the serialized weight string, and consider at most $8T$ frequent candidate pairs when selecting the $T$ leaders.
We impose no additional frequency threshold on the lossless \textsc{Merge} step.
Distortion is measured in quantized-code space relative to the original int8 weights, so the global budget is $\tau^2$ in the notation of Algorithm~\ref{alg:REWRITE}.
We refresh the WeightPE perturbation periodically rather than at every optimization step, using one refresh every ten training updates.
We sweep $\tau_{\mathrm{frac}}\in{0.15,0.20,0.25,0.30}$ for ViT-B/16 and additionally include $0.35$ for ViT-L/16, with $T\in{32,64,128,256}$.

\section{Run environment}
ViT-B/16 experiments are run on an NVIDIA A100-40GB and ViT-L/16 experiments on an NVIDIA H100-80GB GPU.

\section{Run timings}
\label{app:timing}
\begin{figure}[!t]
  \centering
  \includegraphics[width=\linewidth]{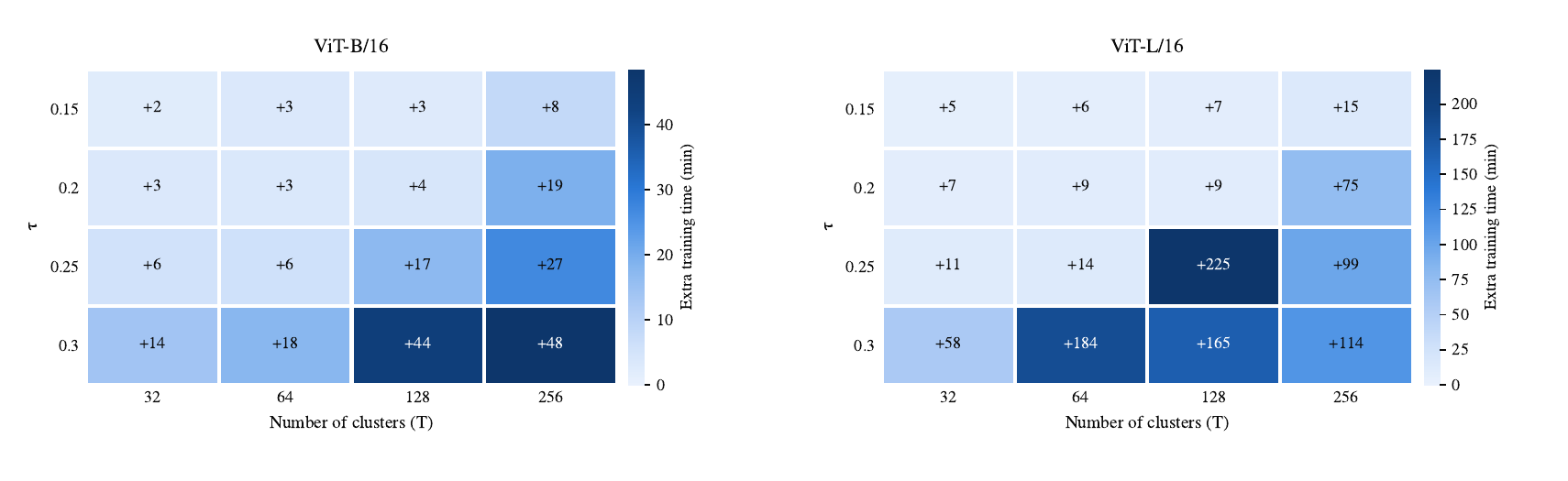}
  \caption{Observed end-to-end training-time overhead across WeightPE sweep grid for ViT-B/16 (left) and ViT-L/16 (right), relative to an int8 QAT baseline.}
  \label{fig:timing-matrix}
\end{figure}

Figure~\ref{fig:timing-matrix} reports the observed wall-clock time of every run in Figures~\ref{fig:tradeoff}, expressed relative to the corresponding $\tau_{\mathrm{frac}}=0$ int8 QAT baseline.

These measurements were collected during normal operation of a shared Slurm-managed cluster. They therefore include uncontrolled system-level variation, such as CPU, I/O, and node contention from co-located jobs, and should not be interpreted as controlled microbenchmarks of the WeightPE operator itself. We report them to provide an indication of the end-to-end training overhead observed in our experimental setting.

\end{document}